\documentclass[journal]{IEEEtran}

\usepackage{cite}
\usepackage{amsmath,amssymb,amsfonts}
\usepackage{graphicx}
\usepackage{textcomp}
\usepackage{xcolor}
\usepackage{booktabs}
\usepackage{multirow}
\usepackage{url}
\usepackage[hidelinks]{hyperref}
\usepackage{orcidlink}
\usepackage{enumitem}

\hypersetup{
  colorlinks=true,
  linkcolor=blue,
  citecolor=blue,
  urlcolor=blue
}

\title{Toward the Whole Picture: Accumulative Fingerprint Mapping and Reconstruction for Small-Area Mobile Sensors}

\author{Xiongjun~Guan$^{\orcidlink{0000-0001-8887-3735}}$,
Jianjiang~Feng$^{\orcidlink{0000-0003-4971-6707}}$,~\IEEEmembership{Member,~IEEE},
and Jie~Zhou$^{\orcidlink{0000-0001-7701-234X}}$,~\IEEEmembership{Senior Member,~IEEE}%
\thanks{Xiongjun Guan, Jianjiang Feng, and Jie Zhou are with the Department of Automation, Tsinghua University, Beijing 100084, China (e-mail: \url{gxj21@mails.tsinghua.edu.cn}; \url{jfeng@tsinghua.edu.cn}; \url{jzhou@tsinghua.edu.cn}).}}

\begin{document}

\maketitle

\begin{abstract}
Small-area fingerprint sensing on mobile devices creates a fundamental mismatch between acquisition and recognition: each touch captures only a tiny, pose-varying local patch, while reliable biometric matching ultimately requires a stable and sufficiently complete fingerprint representation. Existing pipelines largely cope with this mismatch by treating repeated touches as independent partial templates, which leads to repeated registration, repeated matching, and no guarantee of adequate global coverage. In this paper, we advocate a different formulation, namely \emph{accumulative fingerprint mapping and reconstruction} for small-area mobile sensing. Rather than matching every partial patch separately, the proposed perspective converts a sequence of local observations into a unified fingerprint state that is progressively refined as new touches arrive and can be matched only once after consolidation. As a concrete baseline, we present a classical pipeline that performs patch-wise structural feature extraction, feature-level registration and fusion, fingerprint map construction, and phase-based ridge reconstruction. More importantly, we position this baseline within a broader mobile fingerprint framework that integrates structured token learning, two-stage pose reasoning, and diffusion-based generative reconstruction. This viewpoint reframes mobile fingerprint recognition from multi-capture multi-match processing to accumulative map building, state refinement, and one-shot matching, offering a principled route toward efficient, pose-robust, and deployment-friendly biometrics for small-area mobile platforms. The baseline implementation has been publicly released at \url{https://github.com/XiongjunGuan/FpReconstruction}.
\end{abstract}


\section{Introduction}
Small-area fingerprint sensing is becoming a practical default in mobile devices, especially in under-screen sensing and compact touch interfaces. This trend improves industrial integration, but it also exposes a structural weakness in the underlying recognition pipeline: the sensing footprint is often too small to capture a sufficiently informative fingerprint region in a single touch. In practice, each observation reveals only a local patch, while repeated touches vary in location, rotation, pressure, elastic distortion, and overlap. As a result, there is a persistent mismatch between what mobile hardware acquires and what robust biometric recognition actually needs: not isolated local glimpses, but a stable and sufficiently complete fingerprint representation. Classical fingerprint literature has long established the importance of ridge flow, minutiae structure, and template quality for reliable recognition~\cite{maltoni2009handbook}; the mobile setting makes these requirements harder to satisfy, not less relevant.

The standard response is to treat every partial observation as an independent matching unit. Although operationally convenient, this design is fundamentally limited for small-area sensing. It does not guarantee adequate global coverage even after multiple touches; it leaves each local template highly sensitive to pose and contact variation; and it often incurs repeated registration and repeated matching over multiple local templates, increasing computation, latency, and engineering complexity. In other words, conventional pipelines make the user provide more data, but they do not fully convert those additional observations into a better global biometric state.

We argue that the correct abstraction is not repeated partial matching, but \emph{accumulative fingerprint mapping and reconstruction}. Instead of matching every small patch independently, the system should absorb successive touches into a shared fingerprint state, progressively refine that state as new evidence arrives, and perform matching only once on the consolidated result. Under this formulation, repeated touches become complementary observations for map building rather than repeated triggers for downstream comparison. This shift is especially compelling for mobile devices because it aligns algorithm design with natural user behavior: casual repeated touches gradually move the system toward the whole picture of the fingerprint.

The present work contributes at two complementary levels. At the technical level, it provides a reconstruction-oriented classical baseline for this setting. The current repository implements a feature-driven pipeline that extracts orientation and minutiae cues from partial patches, aligns them in a shared spatial frame, fuses them into an accumulative fingerprint map, and reconstructs a larger ridge pattern through phase-based generation. Although deliberately classical and interpretable, this baseline already captures the right systems decomposition of the problem. At the conceptual level, this paper articulates a broader roadmap in which structured patch tokens, two-stage pose reasoning, geometry-aware feature fusion, and generative reconstruction are integrated into a unified mobile fingerprint framework.

The main contributions of this work are summarized as follows.
\begin{itemize}[leftmargin=*]
\item We formulate \emph{accumulative fingerprint mapping and reconstruction} as a new systems paradigm for small-area mobile fingerprint sensing, in which repeated local touches are aggregated into a persistent fingerprint state for one-shot matching rather than treated as independent partial templates.
\item We present a reconstruction-oriented classical baseline, aligned with the current MATLAB prototype, that instantiates this paradigm through patch-wise structural feature extraction, feature-level registration and fusion, map construction, and phase-based fingerprint reconstruction.
\item We introduce a forward-looking framework that explicitly separates initial map building from incremental registration and connects partial fingerprint recognition with structured token learning, pose estimation, dense registration, distortion-aware modeling, image stitching, and diffusion-based generation.
\end{itemize}

\section{Related Work and Perspective}
The proposed formulation lies at the intersection of partial fingerprint recognition, pose estimation, dense registration, distortion-aware modeling, image stitching, and modern generative representation learning. In contrast to conventional related-work taxonomies that isolate these topics, our goal here is to clarify why their combination naturally motivates accumulative reconstruction for mobile fingerprint sensing.

\subsection{Partial Fingerprints and Pose Estimation}
Partial fingerprints are intrinsically difficult because they contain fewer stable correspondences and are more sensitive to local pose variation. Recent work has shown that verification and relative pose alignment are tightly coupled for partial fingerprints, and that solving them jointly can improve both alignment accuracy and identity verification performance~\cite{guan2024joint}. This supports our view that spatial reasoning should be treated as a first-class component in the pipeline rather than a purely auxiliary preprocessing step.

For mobile sensing in particular, absolute pose estimation is also highly relevant. Recent under-screen fingerprint sensing work has demonstrated that explicit finger pose estimation can substantially improve robustness when the sensed area is small and the touch posture varies significantly~\cite{guan2025fingerpose}. This further motivates an accumulative pipeline in which pose information is used to place newly acquired patches into a progressively refined global fingerprint state.

\subsection{Dense Registration and Its Limits}
Pixel-level registration is another important line of work for fingerprint alignment. Recent learning-based methods have improved the quality and efficiency of dense registration by combining phase information and deep feature interaction~\cite{guan2024pdrnet}. However, dense registration is most naturally defined for fingerprint pairs that already share sufficient overlap and image support.

For the mobile partial-patch setting studied here, direct image-level accumulation is more challenging. Fingerprint images are not rigid textures; they are strongly affected by elastic skin distortion, contact variability, and pose changes. Recent work on dense distortion regression has shown that the deformation field of a fingerprint can be substantial and difficult to capture from image evidence alone~\cite{guan2024distortion}. This observation is precisely why we advocate feature-level accumulation before image reconstruction. In our view, direct image-domain accumulation is not the most stable primitive for small-area mobile sensing; the more reliable route is to align and fuse structural features first, and only then synthesize a consolidated fingerprint representation.

\subsection{Structured Tokens and Generative Reconstruction}
Beyond the current classical baseline, our extension path is inspired by two major trends in modern generative modeling. First, vector-quantized autoencoding provides a compelling mechanism for learning discrete, reusable, and compositional structural tokens~\cite{oord2017vqvae}. Such tokens are a natural fit for accumulative fingerprint fusion because they support compact storage, controlled aggregation, and explicit structural regularization.

Second, recent image stitching literature has moved from purely geometric warping toward reconstruction-aware feature synthesis. Unsupervised deep image stitching has shown that stitched features can be reconstructed into coherent images without dense supervision~\cite{nie2021udis}. Implicit neural image stitching further demonstrates that enhanced and blended feature reconstruction can improve seam quality and visual continuity~\cite{zhang2024inis}. Robust stitching research also emphasizes that stitching systems should be resilient under challenging compatibility conditions rather than relying on ideal overlap assumptions~\cite{li2024robuststitch}. These ideas are highly relevant to our setting because they support a general principle: fused intermediate representations can be more reliable than direct pixel-domain composition when geometry is uncertain.

Third, recent diffusion-based methods have demonstrated strong controllability under geometric constraints. In particular, DragDiffusion shows that diffusion features can be manipulated under point-based spatial control while preserving identity and semantic consistency~\cite{shi2023dragdiffusion}. RecDiffusion further shows that diffusion models can be used to correct geometric inconsistency in image stitching and rectangling pipelines~\cite{zhou2024recdiffusion}. Although these works are not designed for fingerprints, their core message is highly relevant: diffusion models can serve not only as image generators, but also as geometry-aware refinement engines. This makes them a promising candidate for the reconstruction stage once fused fingerprint features and relative pose cues are available.

\section{Problem Setting}
Consider a mobile fingerprint sensor that captures a sequence of local patches
\begin{equation}
\mathcal{P}=\{P_i\}_{i=1}^{N},
\end{equation}
where each $P_i$ is a small partial fingerprint observation. Due to natural finger motion and limited sensing area, each patch corresponds to an unknown local region under an unknown relative pose. The objective is not to match every $P_i$ independently. Instead, we seek to estimate a unified latent fingerprint representation
\begin{equation}
\mathcal{R} = \Phi(\mathcal{P}),
\end{equation}
where $\Phi(\cdot)$ denotes the accumulative fusion and reconstruction pipeline.

The reconstructed representation $\mathcal{R}$ should satisfy three properties. First, it should integrate complementary local coverage from multiple observations. Second, it should be more robust to pose variation than any individual patch. Third, it should be suitable for a downstream one-shot matcher, so that repeated captures increase representation quality rather than matching cost.

This problem differs from conventional panorama-style stitching. Fingerprint observations are highly structured, consist of ridge flow and singular local events, and are best fused in a feature space constrained by biometric geometry. It also differs from standard template matching because the target is not a set of independent local templates but a growing fingerprint map and representation that can be reused efficiently.

For clarity, we define the latent state of the system as
\begin{equation}
\mathcal{R} = \{\mathcal{M}, \mathcal{F}, \mathcal{I}\},
\end{equation}
where $\mathcal{M}$ denotes the current fingerprint map, $\mathcal{F}$ denotes the fused structural feature state, and $\mathcal{I}$ denotes the optional reconstructed image-level representation. Under this notation, the full pipeline can be interpreted as state estimation and update over a growing mobile fingerprint map.

\section{Method Overview}
\begin{figure*}[t]
  \centering
  \includegraphics[width=0.96\textwidth]{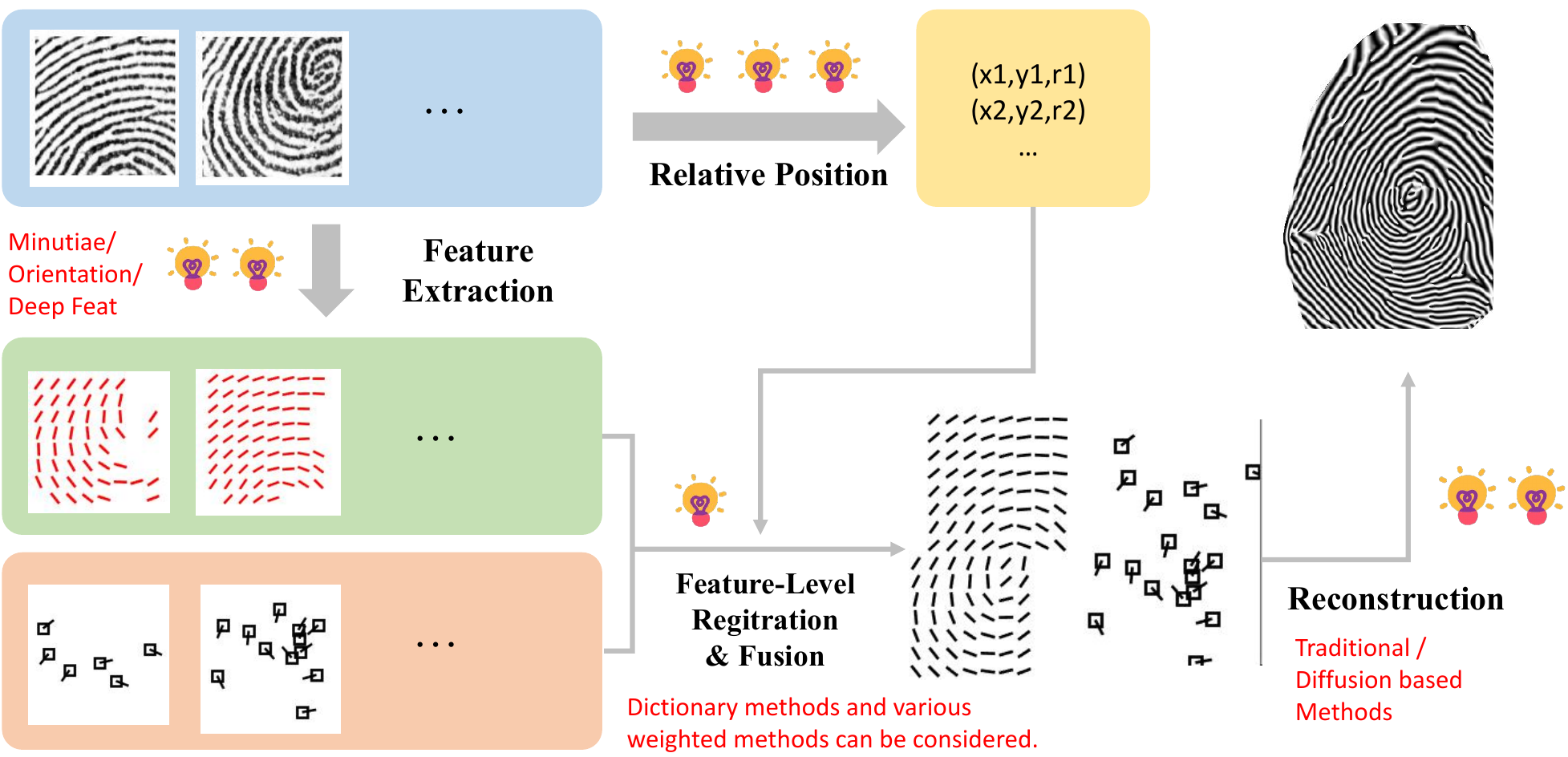}
  \caption{High-level view of the proposed accumulative reconstruction framework. Multiple partial fingerprint patches are converted into structural features, aligned in a common spatial frame, fused across observations, and finally reconstructed into a larger fingerprint representation for one-shot matching. The current repository implements the classical feature-driven baseline of this pipeline. The light-bulb icons in the overview indicate the perceived importance and research headroom of each stage, where more bulbs denote greater development potential and deeper opportunities for future work.}
  \label{fig:overview}
\end{figure*}

Our technical baseline follows the three-stage pipeline shown in Fig.~\ref{fig:overview}: patch-level feature extraction, feature-level registration and fusion, and reconstruction. The current implementation is classical and fully interpretable, while the same structure also supports stronger learning-based variants.

\subsection{Patch-Level Feature Extraction}
For each partial patch $P_i$, the baseline estimates an orientation field and a set of minutiae descriptors. The orientation field provides a dense local description of ridge flow, while minutiae offer sparse but highly discriminative structural anchors. In the current MATLAB prototype, the orientation map is extracted patch-wise and resized into the image resolution. A foreground mask is also estimated to identify valid sensing regions.

Let the feature representation of patch $P_i$ be
\begin{equation}
F_i = \{O_i, M_i, S_i\},
\end{equation}
where $O_i$ denotes orientation information, $M_i$ denotes minutiae, and $S_i$ denotes auxiliary signals such as a foreground mask or confidence support. These features are intentionally classical because they provide high interpretability and strong geometric meaning in partial-overlap settings.

From a systems viewpoint, this stage converts raw sensing observations into map-ready structural evidence. It is the analogue of feature lifting in visual mapping systems: instead of operating directly on raw intensities, later stages operate on a structured representation that is more stable under pose change, local overlap uncertainty, and elastic deformation.

\subsection{Feature-Level Registration and Fusion}
Once features are extracted from multiple patches, we align them into a shared reference space and progressively aggregate them. In the current baseline, this step is realized by explicit relative positioning and region merging. Overlapping regions are fused by averaging compatible orientation evidence and combining minutiae from all available observations.

More generally, we argue that registration in this problem should be understood under a \emph{two-stage pose estimation protocol}. The first stage is \emph{map building}. When no global fingerprint map exists yet, the system must infer the relative arrangement of multiple patches jointly, similar in spirit to structure-from-motion pipelines in classical computer vision. The objective here is not simply pairwise alignment, but the recovery of a globally consistent patch layout:
\begin{equation}
\Pi^{(0)} = \{ \pi_i^{(0)} \}_{i=1}^{N}, \quad \pi_i^{(0)} = (x_i, y_i, \theta_i),
\end{equation}
where the poses are solved together under structural consistency constraints. In this stage, pairwise cues, overlap hypotheses, and graph or optimization-based reasoning are all natural tools.

The second stage is \emph{incremental registration on an existing map}. Once an initial global fingerprint state has been established, each newly incoming patch no longer needs to be jointly optimized with all previous observations from scratch. Instead, it should be localized against the current fingerprint map and used to update the representation:
\begin{equation}
\pi_{t+1} = \Lambda(P_{t+1}, \mathcal{R}_t), \qquad
\mathcal{R}_{t+1} = \Gamma(\mathcal{R}_t, P_{t+1}, \pi_{t+1}),
\end{equation}
where $\Lambda(\cdot)$ denotes localization against the current map, and $\Gamma(\cdot)$ denotes the subsequent fusion update. This distinction is important in practice: map building and incremental registration are related, but they are not the same problem, and a mature mobile system must address both.

Under a unified perspective, both stages can be written as minimizing an energy over pose variables and fused structural consistency:
\begin{equation}
\Pi^\star = \arg\min_{\Pi} \ \mathcal{E}_{\mathrm{pose}}(\Pi;\mathcal{P}) +
\lambda \mathcal{E}_{\mathrm{fusion}}(\Pi,\mathcal{F}),
\end{equation}
where $\Pi$ collects all pose variables, $\mathcal{E}_{\mathrm{pose}}$ penalizes inconsistent spatial layout, and $\mathcal{E}_{\mathrm{fusion}}$ measures disagreement after projecting patch features into the common map frame. In the initial map-building stage, $\Pi$ contains all observed patches and is solved jointly. In the incremental stage, the optimization reduces to the new pose variable conditioned on the existing map state.

Given a set of transformed patch features $\{\widetilde{F}_i\}_{i=1}^{N}$ in a common coordinate frame, the fused representation is written as
\begin{equation}
F^{\star} = \Psi(\widetilde{F}_1,\widetilde{F}_2,\ldots,\widetilde{F}_N),
\end{equation}
where $\Psi(\cdot)$ denotes the accumulation operator. The key point is that fusion happens in a structural feature space instead of directly on raw pixels. This design reduces sensitivity to local illumination, partial contact noise, and pose variation, and it aligns more naturally with the final biometric objective.

In practice, the current repository contains examples for two-patch fusion and multi-patch fusion. Each incoming patch contributes additional ridge-flow evidence and local minutiae support. Thus, the fused representation becomes progressively denser as new captures arrive, which is essential for the accumulative mobile setting.

\subsection{Phase-Based Fingerprint Reconstruction}
After feature fusion, the final objective is to recover a coherent ridge pattern from the fused structural representation. The baseline uses a phase-based reconstruction procedure. Starting from the fused orientation field and minutiae set, it first estimates a valid foreground region, then unwraps the orientation field, and finally synthesizes a fingerprint image through iterative ridge filtering and phase modulation. This design follows the spirit of classical minutiae-to-phase reconstruction methods, which have shown that structural fingerprint cues can effectively constrain dense ridge synthesis~\cite{feng2013fingerprint}.

More concretely, the reconstruction module constructs an initial phase image guided by orientation flow, removes spurious phase singularities, injects spiral phase terms corresponding to minutiae, and generates a final ridge map through
\begin{equation}
I(x,y) = \frac{\cos(\phi(x,y)) + 1}{2},
\end{equation}
where $\phi(x,y)$ is the fused continuous phase after orientation-driven filtering and minutiae-aware correction.

This formulation is well suited to the current baseline for two reasons. First, it leverages classical fingerprint priors directly, making the synthesis process transparent. Second, it translates the fused structural representation into a dense image-like output that can be reused by downstream matching systems without requiring multi-template logic.

\subsection{Unified Objective}
Putting the three stages together, the proposed framework can be understood as optimizing a coupled mobile fingerprint state:
\begin{equation}
\min_{\Pi,\mathcal{F},\mathcal{I}}
\mathcal{E}_{\mathrm{pose}} +
\lambda_1 \mathcal{E}_{\mathrm{feature}} +
\lambda_2 \mathcal{E}_{\mathrm{recon}},
\end{equation}
where $\mathcal{E}_{\mathrm{pose}}$ enforces spatial consistency among partial observations, $\mathcal{E}_{\mathrm{feature}}$ enforces agreement in the fused structural feature space, and $\mathcal{E}_{\mathrm{recon}}$ enforces image-level consistency of the final fingerprint reconstruction. The current repository instantiates this objective through a classical pipeline, while future learning-based variants can replace each term with learned modules and differentiable losses.

\section{Why This Paradigm Matters for Mobile Devices}
The main value of the proposed formulation is system-level rather than cosmetic. For small-area mobile sensors, repeated local captures are inevitable. A conventional pipeline treats them as multiple matching inputs, whereas our formulation treats them as cumulative evidence for building a single stronger representation.

This shift yields several important benefits.
\begin{itemize}[leftmargin=*]
\item \textbf{Coverage improvement.} Different touches often expose different local fingerprint regions. Accumulation naturally improves spatial coverage over time.
\item \textbf{Pose robustness.} Patch alignment and feature-level fusion explicitly absorb translation and rotation variation, reducing sensitivity to touch posture.
\item \textbf{Matching efficiency.} Instead of performing repeated matching over many partial templates, the system can reconstruct a unified fingerprint representation and match it once.
\item \textbf{Deployment simplicity.} One-shot matching simplifies database interaction and downstream recognition logic in practical mobile systems.
\end{itemize}

In short, the proposed pipeline changes the role of repeated captures. They are no longer a source of repeated computation; they become a source of representation refinement.

\section{Practical Mobile Deployment Protocol}
The proposed formulation is especially meaningful when viewed from the perspective of a real mobile biometric system. A practical deployment can be organized around two stages: accumulative enrollment and one-shot verification.

During enrollment, the user provides several natural touches over time. Each new patch is processed into structural features and used to update the current fingerprint state:
\begin{equation}
\mathcal{R}_{t+1} = \Gamma(\mathcal{R}_t, P_{t+1}),
\end{equation}
where $\Gamma(\cdot)$ denotes the incremental fusion and reconstruction operator. In this way, the enrollment template is no longer a single fragile partial print, but a progressively consolidated representation whose quality improves with additional observations.

During verification, the system can either compare a newly accumulated representation against the stored enrollment representation, or directly use the reconstructed fingerprint template for downstream matching. In both cases, the core benefit remains the same: the computational and algorithmic burden is shifted from repeated partial-template matching to representation building. This is an appealing tradeoff for mobile platforms, where robustness and latency are both first-order concerns.

The same protocol also supports flexible user interaction. The user does not need to carefully control patch placement to guarantee full coverage in a single touch. Instead, the system absorbs natural variation over repeated contacts and uses it to refine the fingerprint state. This property is particularly important for small-area sensing, where rigid user cooperation is often unrealistic.

\section{Evaluation Protocol and Future Experiments}
Although this report focuses on formulation and technical design rather than a complete benchmark, the proposed framework naturally suggests a structured evaluation protocol.

\subsection{Stage-wise Evaluation}
We recommend evaluating the system at three levels.
\begin{itemize}[leftmargin=*]
\item \textbf{Pose estimation quality.} Measure both pairwise and global pose accuracy for map-building, as well as localization accuracy for newly added patches on an existing map.
\item \textbf{Reconstruction quality.} Evaluate the completeness, geometric consistency, ridge continuity, and structural realism of the reconstructed fingerprint representation.
\item \textbf{Recognition benefit.} Measure whether accumulative reconstruction improves one-shot matching performance relative to independent partial-template matching.
\end{itemize}

\subsection{Two-Stage Registration Evaluation}
Since we explicitly distinguish map construction from incremental registration, experiments should do the same.

For \emph{map building}, the evaluation should test whether a set of unordered or weakly ordered partial patches can be arranged into a globally consistent fingerprint map. This setting is analogous to multi-view layout recovery in structure-from-motion, except that the underlying scene here is a distorted biometric surface rather than a rigid 3D object.

For \emph{incremental registration}, the evaluation should start from an existing map and test whether a newly arriving patch can be localized and fused correctly. This setting is closer to online localization and map update. In mobile deployment, this stage is particularly important because it determines whether the system can keep refining the user representation over time without repeatedly rebuilding the entire map.

\subsection{Core Comparisons}
The most meaningful comparisons should include:
\begin{itemize}[leftmargin=*]
\item independent partial-template matching without accumulation
\item image-level or dense-registration-first accumulation baselines
\item the current feature-level classical reconstruction baseline
\item future learning-based variants with tokenized features, learned pose estimation, and diffusion-based reconstruction
\end{itemize}

\subsection{Key Questions}
From a scientific perspective, future experiments should answer at least four questions:
\begin{itemize}[leftmargin=*]
\item Does accumulative representation building improve coverage and verification accuracy?
\item Is feature-level accumulation more robust than direct image-domain accumulation under distortion and pose variation?
\item How much of the gain comes from better global map construction versus better incremental localization?
\item How much additional benefit is brought by replacing classical reconstruction with a generative reconstruction prior?
\end{itemize}

\section{Experimental Protocol Draft}
To facilitate future benchmarking and eventual arXiv or journal release, we outline a concrete experimental protocol aligned with the proposed formulation.

\subsection{Tasks}
We recommend organizing experiments into four tasks.
\begin{itemize}[leftmargin=*]
\item \textbf{Patch map construction.} Given a set of partial fingerprint patches without an existing map, estimate a globally consistent patch layout and fused structural state.
\item \textbf{Incremental patch registration.} Given an existing fingerprint map, localize and register a newly incoming patch for state update.
\item \textbf{Fingerprint reconstruction.} Recover a complete or more complete fingerprint representation from fused structural cues.
\item \textbf{One-shot matching.} Evaluate whether the reconstructed or accumulated representation improves final biometric verification relative to non-accumulative baselines.
\end{itemize}

\subsection{Data Organization}
The most natural benchmark unit is not an isolated fingerprint image, but a \emph{capture sequence} associated with one finger. Each sequence should contain multiple partial observations acquired under natural touch variation, including changes in translation, rotation, contact area, and elastic deformation.

For each finger identity, data should preferably be partitioned into:
\begin{itemize}[leftmargin=*]
\item an enrollment sequence used for map construction and accumulative refinement
\item optional intermediate touches used to evaluate incremental registration
\item a verification sequence used for one-shot matching evaluation
\end{itemize}

This organization is important because it matches the real mobile setting. The unit of evaluation is not just whether two local patches match, but whether a user-specific fingerprint state can be built, updated, and reused effectively.

\subsection{Metrics}
We suggest using a stage-aligned evaluation suite.
\begin{itemize}[leftmargin=*]
\item \textbf{Pose metrics.} Relative translation error, rotation error, global layout consistency, and incremental localization accuracy.
\item \textbf{Coverage metrics.} Effective covered area, overlap utilization rate, and accumulated valid-region ratio.
\item \textbf{Reconstruction metrics.} Ridge continuity, structural consistency, orientation agreement, minutiae preservation, and optional image-level measures when reference images are available.
\item \textbf{Recognition metrics.} EER, TAR at specified FAR operating points, DET curves for verification, CMC curves and retrieval curves for identification or search, and matching latency under one-shot evaluation.
\item \textbf{System metrics.} Number of touches needed to reach a target matching accuracy, update cost per newly added patch, and memory footprint of the maintained fingerprint state.
\end{itemize}

\subsection{Baselines}
To properly position the proposed method, comparisons should include:
\begin{itemize}[leftmargin=*]
\item \textbf{Single-patch matching.} Use only one partial fingerprint for enrollment and verification.
\item \textbf{Multi-patch independent matching.} Match multiple partial templates independently and aggregate scores at decision time.
\item \textbf{Image-domain accumulation.} Perform dense-registration-first or stitching-style image accumulation prior to feature extraction.
\item \textbf{Classical feature-level accumulation.} The baseline described in this report.
\item \textbf{Learning-based accumulation.} Future variants using structured tokenization, learned pose estimation, and diffusion-based reconstruction.
\end{itemize}

\subsection{Ablation Design}
The ablation study should directly mirror the conceptual decomposition of the paper.
\begin{itemize}[leftmargin=*]
\item remove feature-level fusion and compare against direct image accumulation
\item remove global map construction and keep only incremental local registration
\item remove incremental localization and rebuild the map from scratch each time
\item replace classical reconstruction with learned reconstruction while keeping the same fused feature state
\item vary the number of accumulated touches and measure the accuracy-efficiency tradeoff
\end{itemize}

\subsection{Expected Claims}
If the proposed formulation is correct, the experimental evidence should support the following claims:
\begin{itemize}[leftmargin=*]
\item accumulative representation building improves effective fingerprint coverage
\item one-shot matching on the accumulated state is more efficient than repeated matching over multiple local templates
\item feature-level accumulation is more robust than direct image-domain accumulation under distortion and pose change
\item separating map construction from incremental registration is beneficial for practical mobile deployment
\item stronger learned reconstruction modules can further improve the baseline without changing the core formulation
\end{itemize}

\section{Expected Results and Discussion}
Although this report does not yet include a finalized benchmark table, the proposed framework leads to several concrete expectations about experimental outcomes.

\subsection{Expected Empirical Trends}
First, we expect fingerprint coverage to improve monotonically, or near-monotonically, as more partial touches are accumulated. This is the most immediate consequence of the proposed formulation. Unlike independent partial-template matching, where each new touch is evaluated largely in isolation, the proposed system converts additional touches into a stronger shared fingerprint state.

Second, we expect the main performance gains to appear under realistic mobile variation rather than under ideal overlap conditions. In easy cases with strong overlap and low distortion, direct matching of partial prints may already be sufficient. The value of accumulative reconstruction should become most visible when touches are spatially incomplete, pose-varying, or moderately distorted, which is precisely the regime that motivates the problem.

Third, we expect feature-level accumulation to outperform direct image-domain accumulation in the presence of deformation. This expectation is consistent with the structural nature of fingerprint data and with prior evidence that dense distortion can significantly affect image-level registration quality. If this trend is confirmed experimentally, it would directly support the central design choice of the paper.

Fourth, we expect one-shot matching on the accumulated fingerprint state to provide a better efficiency-accuracy tradeoff than repeated matching over multiple independent local templates. This claim is especially important from a deployment perspective because it links the method not only to recognition accuracy, but also to system simplicity and runtime efficiency.

\subsection{Interpreting Positive Results}
If the empirical evidence follows these trends, the significance of the paper would go beyond improving a particular baseline. It would suggest that small-area mobile fingerprint recognition is better understood as a \emph{state-building problem} than as a sequence of independent pairwise matching problems. In that case, map construction, incremental localization, feature fusion, and reconstruction become native parts of the biometric pipeline rather than optional enhancements.

This would also strengthen the role of the present baseline. Even if the current implementation is classical, strong positive results would indicate that the formulation itself is correct, and that more advanced learning-based modules can be developed on top of a sound problem decomposition.

\subsection{Interpreting Negative or Mixed Results}
Negative or mixed results would still be informative. If accumulative reconstruction improves coverage but not verification performance, this may indicate that the fused representation is not yet sufficiently discriminative, or that the reconstruction stage introduces artifacts that weaken matching. If global map construction works but incremental registration remains unstable, this would suggest that localization against an evolving fingerprint state is itself a major research challenge. If image-domain accumulation performs similarly to feature-level accumulation in some settings, then the operating regime of the method should be characterized more carefully rather than assumed universally.

From a research perspective, these outcomes would not invalidate the formulation. Instead, they would clarify which components deserve the most attention: pose estimation, structural tokenization, reconstruction fidelity, or downstream matcher compatibility.

\section{Current Baseline in the Repository}
The current repository should be understood as a reconstruction-oriented baseline rather than a fully learned end-to-end system. It is rooted in traditional fingerprint processing and is designed to be interpretable, modular, and extensible.

The implemented code currently covers the following capabilities.
\begin{itemize}[leftmargin=*]
\item Orientation estimation from local patch observations.
\item Minutiae loading and aggregation across multiple patches.
\item Explicit feature-level merging under known or estimated relative placement.
\item Phase-based reconstruction driven by fused orientation and minutiae cues.
\end{itemize}

This baseline is important because it establishes a credible lower-bound system for the mobile accumulative setting. It makes the key design choice explicit: the right abstraction is to fuse partial observations into a unified structural representation before reconstruction and matching.

At the same time, the current baseline is not the final intended system. It does not yet fully exploit learned feature dictionaries, confidence-aware registration, or modern generative priors. Those directions are the natural next step.

\section{A Forward-Looking Learning-Based Framework}
\begin{figure*}[t]
  \centering
  \includegraphics[width=0.96\textwidth]{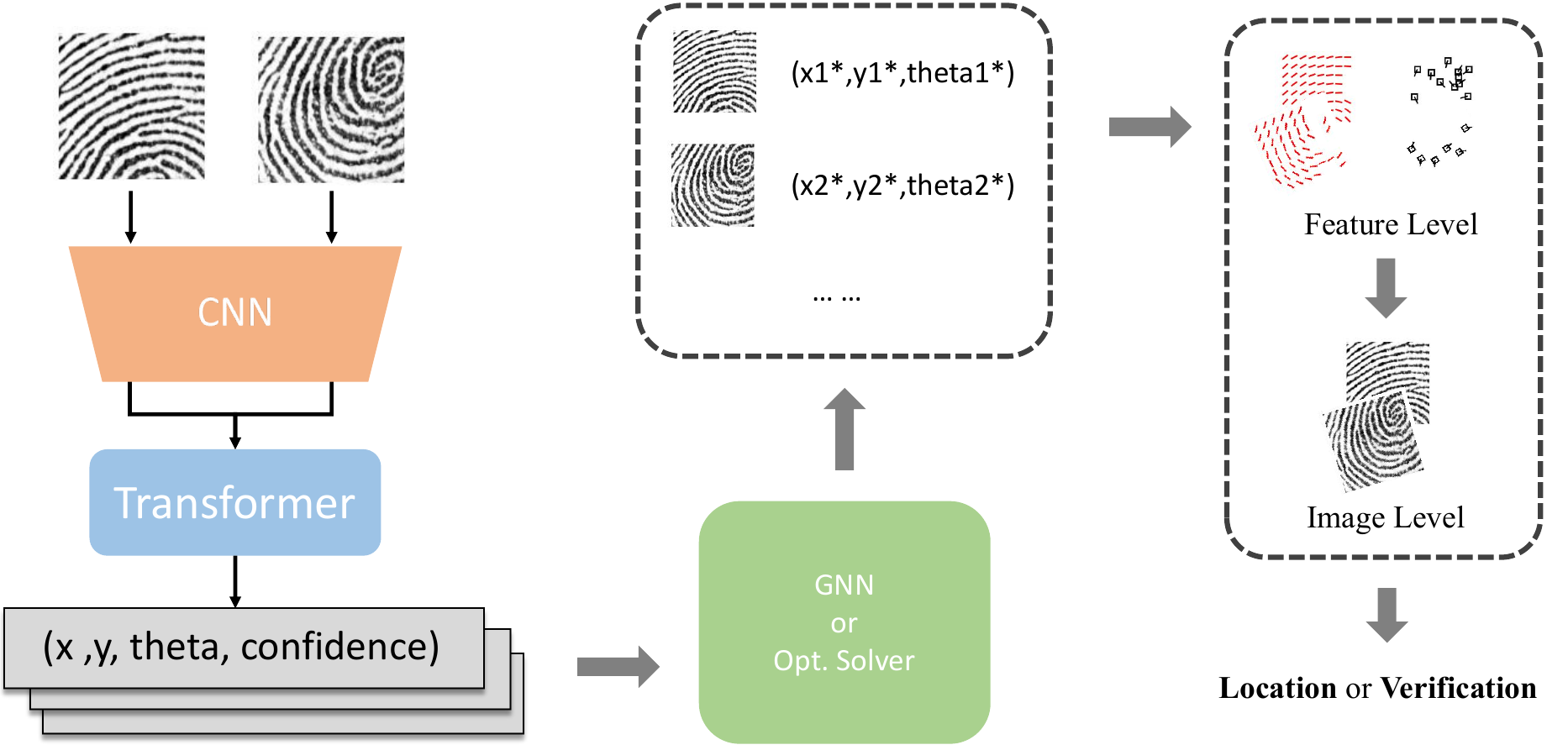}
  \caption{Forward-looking framework beyond the current baseline. Partial patches are encoded into structured patch representations, associated with relative pose variables, fused through graph or optimization reasoning, and finally reconstructed at the image level. This figure represents the intended extension path rather than the fully implemented functionality of the current repository.}
  \label{fig:framework}
\end{figure*}

Fig.~\ref{fig:framework} illustrates the broader framework that motivates the long-term evolution of this line of work.

\subsection{Structured Feature Learning}
The first extension is to replace hand-crafted or purely local features with learned structured representations. A promising direction is to use a dictionary-style tokenizer inspired by vector-quantized autoencoding~\cite{oord2017vqvae}. Under this view, each partial patch is mapped to a set of discrete or semi-discrete structural codes that capture ridge flow, local minutiae context, and regional texture quality.

This strategy is appealing because it balances representation compactness and interpretability. Instead of treating a patch as an unconstrained feature vector, the model learns a compositional vocabulary of fingerprint structures. Such tokens are easier to accumulate, compare, and regularize during multi-patch fusion.

\subsection{Relative Pose Estimation and Fusion}
The second extension is to predict relative spatial relationships among patches jointly with their structural features. This can be done through pairwise pose estimation~\cite{guan2024joint}, graph neural reasoning~\cite{kipf2017gcn}, transformer-style relational modeling~\cite{dosovitskiy2021vit}, or explicit optimization over patch coordinates and rotations. We emphasize again that this module should serve both global map building and incremental registration on an existing map.

Formally, let each patch be associated with a pose variable
\begin{equation}
\pi_i = (x_i, y_i, \theta_i, c_i),
\end{equation}
where $(x_i,y_i,\theta_i)$ denote relative placement and $c_i$ denotes confidence. Fusion becomes a confidence-aware structured aggregation problem:
\begin{equation}
F^{\star} = \Psi\big(\{(F_i,\pi_i)\}_{i=1}^{N}\big).
\end{equation}
At the map-building stage, the set of poses should be estimated jointly under global consistency. At the incremental stage, the new pose variable should be estimated relative to the current map state rather than from scratch. This formulation is especially suitable for mobile sensing because the sequence of touches is inherently incremental and pose-variant.

\subsection{Generative Reconstruction}
The third extension is to upgrade reconstruction from a classical phase-based module to a stronger generative prior. Diffusion models~\cite{ho2020ddpm,shi2023dragdiffusion,zhou2024recdiffusion} are particularly attractive because they can recover globally consistent ridge structures while respecting local feature constraints. In such a system, fused structural features would act as conditioning signals, and the final generator would synthesize a complete high-fidelity fingerprint image or latent biometric template.

Importantly, this extension does not invalidate the baseline; it generalizes it. The current repository already demonstrates the correct decomposition of the problem into feature extraction, fusion, and reconstruction. A learning-based generator would simply replace the last stage with a more expressive conditional prior.

\section{Discussion}
From a high-level perspective, we believe the most important message of this report is not a specific reconstruction formula, but a systems thesis: \emph{small-area mobile fingerprint sensing should move from multi-capture multi-match processing to accumulative fingerprint mapping, reconstruction, and one-shot matching.}

This thesis has several implications. First, it suggests that representation quality should be improved online as the user naturally interacts with the sensor. Second, it elevates pose estimation and structural fusion to first-class modules in the biometric pipeline. Third, it creates a natural bridge between classical fingerprint geometry and modern generative modeling.

The present baseline is intentionally conservative and transparent. This makes it a useful reference point for future work, including learned patch tokenization, confidence-driven fusion, graph-based spatial reasoning, and conditional diffusion reconstruction. It also helps separate what is already demonstrated from what remains to be built.

Naturally, the current technical report has limitations. The repository baseline is not yet a fully self-contained end-to-end benchmark system, and several implementation dependencies remain outside the public snapshot. In addition, quantitative evaluation and large-scale matching experiments are beyond the scope of this report. Nevertheless, the formulation, code structure, and pipeline decomposition already provide a strong foundation for future mobile biometric research.

\section{Limitations}
This work has several limitations that should be stated explicitly.

First, the current repository represents a reconstruction-oriented baseline rather than a polished production system. It captures the core pipeline and the intended problem decomposition, but it does not yet provide the full set of engineering components needed for large-scale benchmark release.

Second, the present technical report emphasizes formulation and system design more than exhaustive empirical validation. While we have outlined a rigorous evaluation protocol, a complete set of quantitative experiments remains future work.

Third, the current classical baseline does not yet exploit learned structural tokenization, confidence-aware map update, or generative reconstruction priors. These elements are expected to be important if the framework is scaled toward stronger performance.

Fourth, the mobile fingerprint setting itself remains challenging. Patch overlap may be weak, contact distortion may be severe, and user behavior may be inconsistent across sessions. Any practical deployment will therefore depend not only on algorithm design, but also on sensing quality and data collection protocol.

\section{Broader Impact}
The proposed formulation is motivated by practical mobile biometric sensing, where limited sensor area creates a real tension between usability, efficiency, and recognition reliability. A successful accumulative reconstruction framework could reduce the number of failed matches, improve the utility of small sensors, and simplify downstream matching pipelines by converting repeated local captures into one-shot comparison.

At the same time, any improvement in biometric sensing should be considered carefully. More efficient or more robust fingerprint recognition can be beneficial for secure device access, but it also increases the importance of responsible handling of biometric data, secure on-device storage, and clear user consent. The proposed formulation should therefore be understood as a technical contribution to sensing and representation, not as an argument for broader biometric deployment without appropriate safeguards.

\section{Conclusion}
This technical report argues that small-area mobile fingerprint recognition should be rethought as an accumulative mapping and reconstruction problem rather than a repeated partial-matching problem. Instead of comparing many incomplete local captures independently, the proposed paradigm progressively consolidates them into a stronger fingerprint state that moves the system toward the whole picture of the underlying finger and enables one-shot matching after accumulation. The current repository provides a classical baseline built on structural feature extraction, feature-level fusion, map construction, and phase-based reconstruction, while the broader framework points toward structured token learning, two-stage pose reasoning, and diffusion-based generation. We believe this direction is well aligned with the realities of mobile sensing and offers a promising path toward efficient, robust, and deployment-friendly fingerprint biometrics.

\bibliographystyle{IEEEtran}
\bibliography{refs}

\end{document}